\documentclass[letterpaper, 10 pt, conference]{ieeeconf}  

\IEEEoverridecommandlockouts                              

\usepackage{times} 
\usepackage{multicol}
\usepackage{lipsum}
\usepackage[bookmarks=true]{hyperref}
\usepackage{xcolor}
\usepackage{graphicx}
\usepackage{subfigure}
\usepackage{float}
\usepackage{booktabs}
\usepackage{caption}
\usepackage{multicol}
\usepackage{amsmath}
\usepackage[T1]{fontenc}

\newcommand{\dash}{\textsc{DASH}}
\newcommand{\dashvone}{\textsc{DASH-v1}}
\newcommand{\dashvtwo}{\textsc{DASH-v2}}
\newcommand{\dashvthree}{\textsc{DASH-v3}}
\newcommand{\dashvfour}{\textsc{DASH-v4}}
\newcommand{\dashvfive}{\textsc{DASH-v5}}

\newcommand{\vone}{\textsc{v1}}
\newcommand{\vtwo}{\textsc{v2}}
\newcommand{\vthree}{\textsc{v3}}
\newcommand{\vfour}{\textsc{v4}}
\newcommand{\vfive}{\textsc{v5}}

\begin{document}


\title{\LARGE \bf Designing Anthropomorphic Soft Hands through Interaction}

\author{
  \authorblockN{Pragna Mannam\authorrefmark{1}, Kenneth Shaw\authorrefmark{1}, Dominik Bauer, Jean Oh, Deepak Pathak,
    and Nancy Pollard}
  \authorblockA{Robotics Institute, Carnegie Mellon
    University \\
    \authorrefmark{1} Equal Contribution \\
  \texttt{pmannam@andrew.cmu.edu}}
}

\maketitle
\thispagestyle{empty}
\pagestyle{empty}


\begin{abstract}
Modeling and simulating soft robot hands can aid in design iteration for complex and high degree-of-freedom (DoF) morphologies. This can be further supplemented by iterating on the design based on its performance in real world manipulation tasks. However, iterating in the real world requires an approach that allows us to test new designs quickly at low costs. In this paper, we leverage rapid prototyping of the hand using 3D-printing, and utilize teleoperation to evaluate the hand in real world manipulation tasks.  Using this method, we design a 3D-printed 16-DoF dexterous anthropomorphic soft hand (DASH) and iteratively improve its design over five iterations. Rapid prototyping techniques such as 3D-printing allow us to directly evaluate the fabricated hand without modeling it in simulation. We show that the design improves over five design iterations through evaluating the hand’s performance in 30 real-world teleoperated manipulation tasks. Testing over 900 demonstrations shows that our final version of DASH can solve 19 of the 30 tasks compared to Allegro, a popular rigid hand in the market, which can only solve 7 tasks. We open-source our CAD models as well as the teleoperated dataset for further study. They are made available on our website \url{https://dash-through-interaction.github.io}.
\end{abstract}


\section{Introduction}
\label{sec:Introduction}
Rapid prototyping technologies have advanced significantly, making way for designers to build new systems at a fast pace. These techniques, such as 3D-printing, allow for quick turnaround between design iterations to test and evaluate systems quickly. This is especially useful for systems with dynamics that are difficult to predict or model, such as soft robot manipulators. 

Iterating for dexterous soft hand designs is a laborious process. The complex design space and the infinite degrees of freedom make it difficult to predict the effects of incremental design changes. Unlike rigid robot hands, state-of-the-art soft body simulators are not able to provide accurate, efficient, and robust evaluation of soft designs~\cite{chen2020design}. 
Hands such as the BRL/Pisa/IIT SoftHand~\cite{della2018toward} or the RBO hand~\cite{Puhlmann_2022} have evolved over years to incorporate more adaptive synergies and dexterity. 
To speed up development times and reduce fabrication overhead many works have recently turned towards 3D-printing to either directly print soft hands~\cite{bauer2022towards} or to quickly create complex molds~\cite{zhang2022creation}. 
While this has significantly reduced the cycle time for fabrication, designing dexterous soft hands still requires a lot of expertise, and trial and error due to the continuously deformable nature of soft robots. The lack of appropriate simulators means the evaluation of soft hand designs has to be done on the real prototype by using hand-crafted policies~\cite{abondance2020dexterous} or sequential keyframed open-loop poses~\cite{bauer2022towards}.


\begin{figure}[t]
    \centering
    \includegraphics[width=\linewidth]{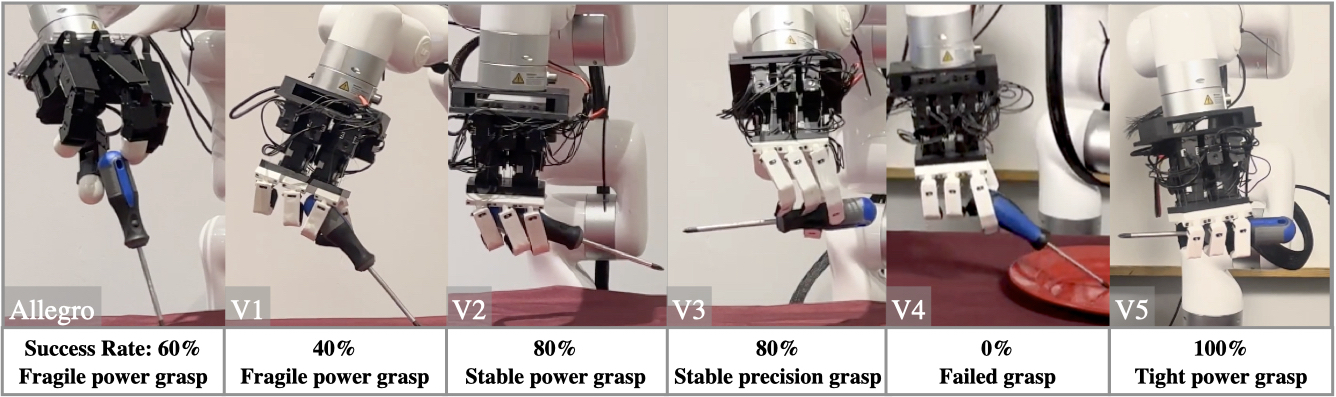}
    \caption{
    Manipulation task performance over five iterations of \dash{} designed through rapid prototyping and real-world evaluation on tasks alongside task performance of our baseline hand Allegro.
    }
    \vspace{-.7cm}
    \label{fig:first_fig}
\end{figure}

\begin{figure*}[th!]
 \centering
 \includegraphics[width=0.9\linewidth]{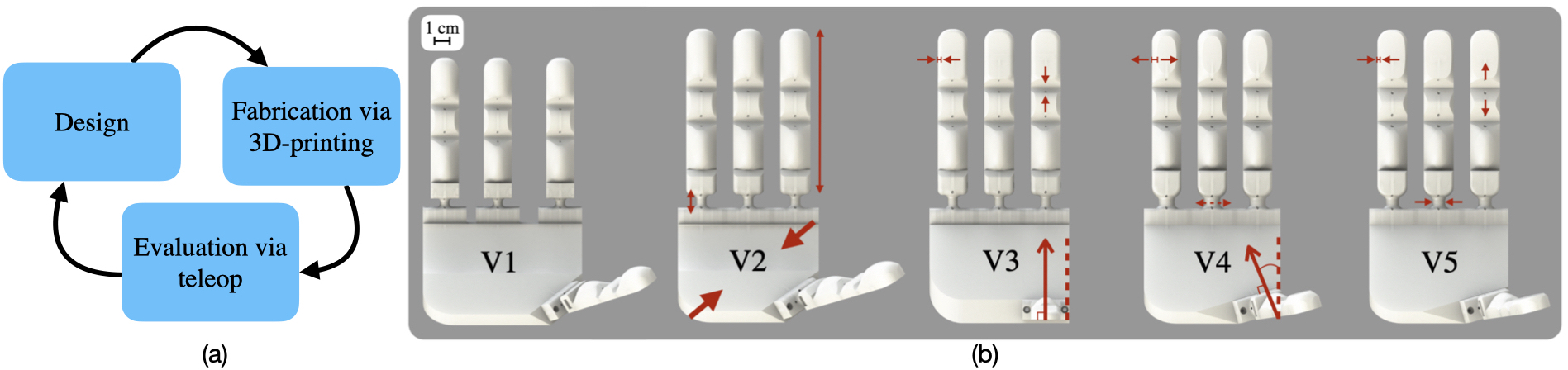} \captionof{figure}{(a) Our soft robotic hand design process involving rapid prototyping and real-world evaluation (b) CAD models and differences across \dash{} iterations \vone{} through \vfive{}, as explained in Section~\ref{sec:DASHdesignandexpts}.}
 \vspace{-.7cm}
 \label{fig:framework_and_CAD}
\end{figure*}

Our key insight is that we can evaluate these systems beyond hand-crafted policies or sequential keyframed open-loop poses using recent advancements in teleoperation systems. Improvements in hand tracking and pose estimation \cite{FrankMocap_2021_ICCV} have led to the development of vision-based teleoperation approaches including using a single RGB camera for real-time tracking of human hand poses~\cite{sivakumar2022robotic}. Teleoperation offers valuable insights into system performance and enables the identification of robust strategies in real-world scenarios. Simulation often falls short in capturing system nuances, accurately modeling soft materials, and adapting strategies. Therefore, tasks that succeed in simulation can still fail when observed and evaluated through teleoperation, providing a clearer understanding of the system's capabilities and limitations.


In this paper, we 3D-print soft robotic hands to test iteratively using teleoperation on a designated manipulation task set, modify the design, and repeat this process shown in Figure~\ref{fig:framework_and_CAD}(a). While manually testing and revising designs to improve systems is not a new concept~\cite{zhang2022creation}, recent technologies allow us to repeatedly iterate the entire framework of designing, fabricating, and evaluating in a matter of \textit{days}. This framework is versatile and can be implemented at any stage of the design process. Its purpose is to bridge the gap between the real world and simulation by adjusting for discrepancies or expediting fine-tuning for real-world scenarios. We envision this framework as a valuable augmentation to existing design frameworks, enhancing the overall design process for soft anthropomorphic robotic hands.

Using our framework, we present a case study to design a $16$-DoF tendon-driven 3D-printed soft hand \dash{}, shown in Figure~\ref{fig:first_fig}. 
This hand has a small form factor similar in size to a human hand, 3D-printable parts that are easily replaceable, and a modular customizable design that allows for easy iteration. 
Through teleoperation, we explore the capabilities of the soft hand in order to inform our design iterations across five hands: \dash{} \vone{}, \vtwo{}, \vthree{}, \vfour{}, and \vfive{}.
In order to evaluate the dexterity of the hand, we designed a suite of $30$ manipulation tasks with varying grasp types and objects that are inspired by human hand capabilities, which allows us to test the capabilities of our robot hand. 
Our hands show improvements across iterations, albeit not monotonically, and each iteration, except \vone{}, required less than $100$ hours to design, fabricate, and test.
\dash{} \vone{}, \vtwo{}, \vthree{}, \vfour{}, and \vfive{} succeed on $70\%$, $82\%$, $83\%$, $75\%$, and $87\%$ of executions across all tasks, respectively. We also outperform a commercial dexterous robotic hand, Allegro~\cite{lee2016kitech}, which has a success rate of $60\%$ on the same $30$ tasks.

The contributions of this paper are
\begin{itemize}
    \item A detailed report of our process for designing soft hands that leverages rapid prototyping techniques and uses a teleoperated real robot for evaluation, instead of simulation.
   \item The design of a state-of-the-art dexterous anthropomorphic soft hand using our framework, that outperforms a commercial robotic hand on real world manipulation tasks.
    \item The release of open-source CAD models and data corresponding to $900$ teleoperated human demonstrations to democratize access to low-cost dexterous hands.
\end{itemize}

\vspace{-0.4cm}
\section{Related Work}
\label{sec:RelatedWork}

Soft robotic hands such as RBO~\cite{Puhlmann_2022} utilize intrinsic compliance, rapid prototyping, and actuated palms for a modular, highly compliant, high degree-of-freedom, and low cost manipulator in order to perform a large variety of in-hand manipulation tasks and object grasps. However, existing robotic hands are still far from achieving human-like dexterity for manipulation~\cite{Controzzi2014}. It is necessary to continuously improve and refine the design of both existing and new robotic hands in order to achieve greater dexterity and functionality for performing more complex manipulation tasks.

Although there is currently no unified framework for designing iterations of soft anthropomorphic hands, design iteration methods for robotic systems have been explored. Typically, Finite Element Method (FEM) is used to assess optimal geometries and morphologies before fabricating the final design for real-world evaluation \cite{feng2018soft,elsayed2014finite}. For instance, the SOFA soft robot simulator has been used to co-optimize control and design of soft hands \cite{deimel2017automated}. However, these approaches are primarily limited to simulated environments and do not address the sim-to-real gap or real-world design iteration. A related framework in robotic fish design utilizes simulation-based FEM testing, real-world design iteration, and proposes a modular design for easier iterations \cite{zhang2022creation}. Nevertheless, optimizing soft robots is challenging due to their complex geometries, impeding the development of efficient optimization algorithms. Furthermore, the lack of efficient simulation tools that can rapidly evaluate design candidates further compounds the challenge \cite{chen2020design}.

Learning control policies for dexterous manipulation is challenging due to the high DoFs and complex interactions~\cite{nagabandi2020deep,andrychowicz2020learning}. In contrast, teleoperation offers a swift and natural way to control robot hands, beyond pick-and-place scenarios. It has been used for human demonstrations in imitation learning~\cite{dime_lerrel, videodex}. Teleoperation is particularly valuable during the design process of dexterous hands, enabling quick evaluation of nuanced capabilities. Mapping human to robot hand morphology can be categorized as \textit{Joint-to-Joint}, \textit{Point-to-Point}, or \textit{Pose-based}~\cite{li2022survey}. For our soft hand, we adopt a similar joint-to-joint mapping technique as Liarokapis et al.~\cite{liarokapis2013telemanipulation}.
 
Our work extends the design and fabrication methodology presented by Bauer et al.~\cite{bauer2022towards}, consisting of simplifying the design complexity of soft hands by incorporating geometric features such as bumps or creases to achieve `joint-like' deformations. They perform kinematic testing on designs before fabricating and evaluate a single design. Similar to Bauer et al.~\cite{bauer2022towards}, we utilize 3D printing, creases for `joint-like' deformations, and tendon-driven actuation to curl the soft fingers. In addition to these features, \dash{}'s design incorporates three additional tendons in all fingers, enabling adduction, abduction, and folding of the fingers towards the palm, thereby enhancing dexterity.

\section{Experiment Setup}
\label{sec:Setup}
\subsection{Robot Hand design}

\label{sec:setup_hand}
\subsubsection{Finger Joints}
All iterations of \dash{} consist of four fingers: the thumb, index, middle, and ring fingers (see Figure~\ref{fig:assembly_calibration}(a)). 
In order to achieve modularity, each finger, including the thumb, is designed identically. 
Each finger has three joints (from the base of the finger to the fingertip): the metacarpophalangeal (MCP) joint,  proximal interphalangeal (PIP) joint, and the distal interphalangeal (DIP) joint. The joints for each finger are shown in Figure~\ref{fig:assembly_calibration}(b).

\subsubsection{Tendons}
\begin{figure}
    \centering
    \includegraphics[width=\linewidth]{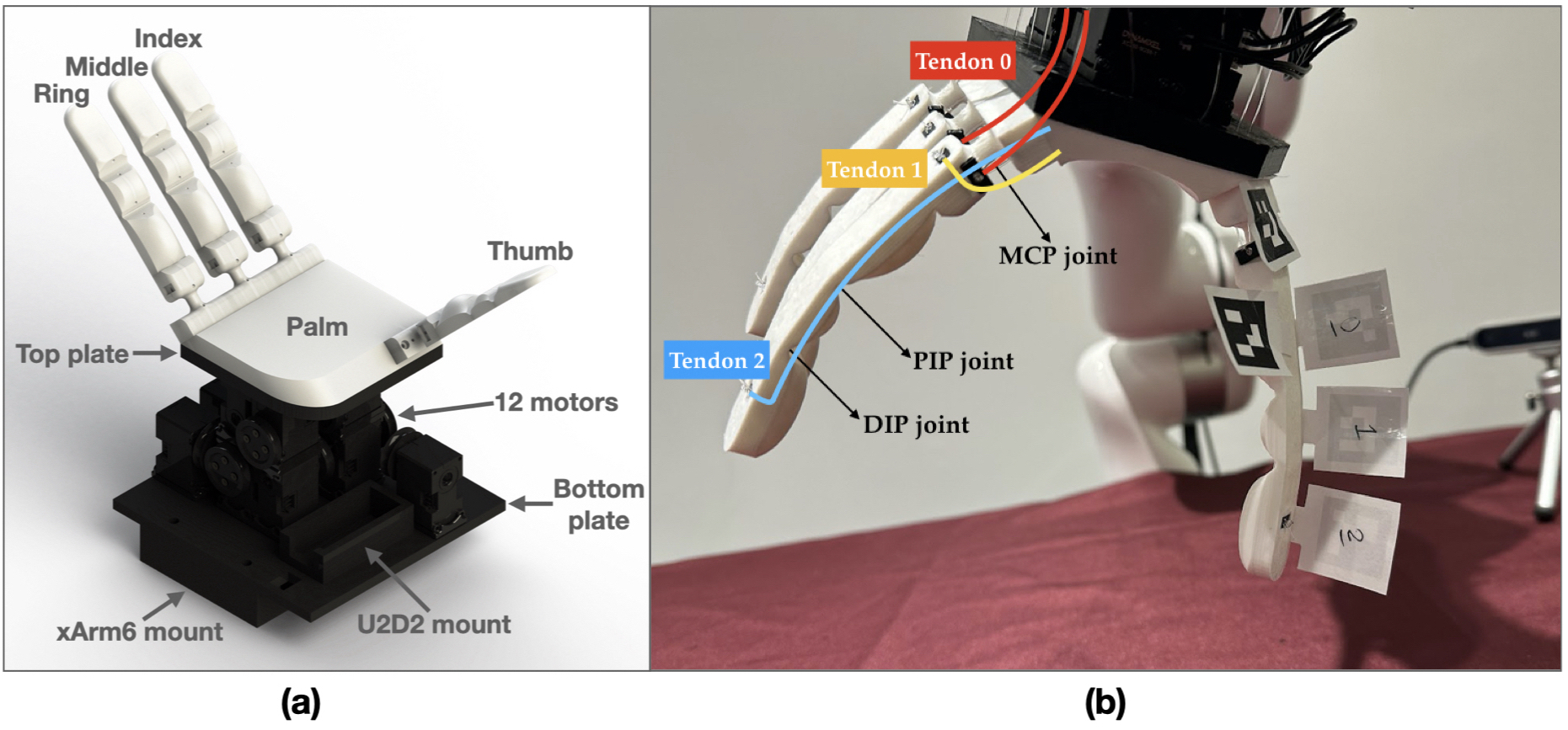}
    \caption{(a) Assembly of \dashvthree{} (top to bottom) including fingers, palm, top plate, motors, bottom plate, xArm6 mount. (b) Calibration procedure to map motor angles to finger joint positions, where tendon 0 actuates MCP side-to-side, tendon 1 actuates MCP forward folding motion, and tendon 2 curls the finger controlling both PIP and DIP joints.}
    \vspace{-.7cm}
    \label{fig:assembly_calibration}
\end{figure}

Each finger is controlled by four tendons shown in Figure~\ref{fig:assembly_calibration}(b).
Two tendons run along the sides of the MCP joint, closest to the palm, for abduction and adduction, which allows the fingers to move closer together and farther apart. These two tendons are controlled by a single motor, so we refer to them as tendon 0. 
A single tendon, tendon 1, is used to flex the finger forward at the MCP joint, orthogonally to the axis of motion of the abduction-adduction tendons.
The last tendon, tendon 2, runs through the entire length of the finger to enable completely curling into itself. 

\subsection{Fabrication using 3D-printing}
The hand assembly, shown in Figure~\ref{fig:assembly_calibration}(a), is the same for all hand iterations and consists of 4 soft fingers attached to the soft palm. The rigid components include a top plate below the palm, 12 motors, a bottom plate which also houses the Dynamixel U2D2 motor controller, and a xArm6 mount.

The soft dexterous hand's mounts, motor housing, and motor pulleys are all 3D-printed from PLA (rigid material depicted in black in Figure~\ref{fig:assembly_calibration}(a)) while the soft hand was printed with Ninjaflex Edge (83A shore hardness)~\cite{ninjaflex} using an Ender 3 S1 Plus. 
For all the robot experiments, the hand was mounted onto a xArm6~\cite{xarm6} robot arm. DASH costs approximately \$$1500$ to build with the majority of the cost consisting of 3D-printer (\$$500$) and the twelve motors (\$$1000$) required. For comparison, the Allegro hand is also 16-DoF and costs around \$$15000$ \cite{zhu2019dexterous}. 

\subsection{Hand Evaluation using Teleoperation}
\label{sec:setup_control}


\subsubsection{Learning Kinematics}\label{sec:kinematics} To approximate the kinematics of the finger joints without real-time feedback, we learn a model from collecting offline paired motor and joint angle data from a single finger. 
Through small increments of 3 degrees of actuation and joint angle tracking across $1000$ finger configurations using AR tags and RGB cameras, we obtain a collection of tuples (joint angles, motor angles) that are normalized to $[0, 1]$ for independent kinematic calibration, assuming fixed joint lengths and bending at the creases.

A linear model is learned using the collected data to map finger joint angles to motor angle outputs. We refer to the motors controlling tendons 0, 1, and 2, as shown in Figure~\ref{fig:assembly_calibration}(b), as motors 0, 1, and 2, respectively. The equations for MCP, PIP, and DIP joint angles are shown below. In equation~\ref{eq:MCP}, we learn the MCP joint angles $\theta_{\mathsf{MCP}_{\mathsf{side}}}, \theta_{\mathsf{MCP}_{\mathsf{fwd}}}$ jointly since the amount of side-to-side angle at the MCP joint can restrict the forward folding motion of the finger. In equation~\ref{eq:DIP_PIP}, the Motor 2 angle $\theta_{\mathsf{motor}_2}$ is an average measure of the motor angle for the desired PIP and DIP joint angles $\theta_{\mathsf{PIP}}, \theta_{\mathsf{DIP}}$ since the same tendon controls both the PIP and DIP joints. The weights in equations~\ref{eq:MCP} and~\ref{eq:DIP_PIP} are found by fitting our data using linear functions. 
We collect training data for almost two hours for each iteration of the hand to calibrate new models, using the weights shown in Table~\ref{tab:calibration_weights}.

\iftrue
\begin{align}\label{eq:MCP}
    \begin{bmatrix}
    \theta_{\mathsf{motor}_0} \\ \theta_{\mathsf{motor}_1}
    \end{bmatrix} &= 
    \begin{bmatrix}
        w_1 & w_3 \\ w_2 & w_4
    \end{bmatrix} \cdot \begin{bmatrix}
        \theta_{\mathsf{MCP}_{\mathsf{side}}} \\ \theta_{\mathsf{MCP}_{\mathsf{fwd}}}
    \end{bmatrix}
    + \begin{bmatrix}
        b_1 \\b_2
    \end{bmatrix}
\end{align}

\begin{equation} \label{eq:DIP_PIP}
     \theta_{\mathsf{motor}_2} = 
   \displaystyle \frac{\theta_\mathsf{PIP} w_5 + b_3}{2} + \frac{\theta_\mathsf{DIP} w_6 + b_4}{2}
\end{equation}
\fi

\begin{table}[t]
\centering
\small
\begin{tabular}{@{} *7l @{}}    \toprule
\emph{\textbf{Hand Design}} & \emph{v1 } & \emph{v2} & \emph{v3} & \emph{v4} & \emph{v5}  \\ \midrule
$w_1$ & $-1.05$  & $-0.43$  &  $-0.43$ & $-0.59$ & $-0.59$ \\
$w_2$ &  $0.01$ &  $0.2$ &  $0.2$ & $-0.12$ & $-0.19$ \\
$w_3$ & $0.1$ &  $0.51$ &  $0.51$ & $0.26$  & $-0.32$ \\
$w_4$ & $0.83$  &  $0.54$ & $0.54$ & $0.38$ & $0.72$ \\
$w_5$ & $0.67$  &  $0.6$ &  $0.6$ & $0.62$ & $0.63$ \\
$w_6$ &  $0.99$ & $0.76$  &  $0.76$ & $1.69$ & $0.65$ \\
$b_1$ & $0.47$  & $0.38$  &  $0.38$ & $0.45$ & $0.58$ \\
$b_2$ & $-0.07$  & $0.01$  & $0.01$ & $0.44$ & $-0.03$  \\
$b_3$ &  $0.03$ &  $-0.04$ & $-0.04$ & $-0.05$ & $-0.09$  \\
$b_4$ & $-0.01$  & $-0.16$  &  $-0.16$ & $-0.3$ & $-0.07$ \\
\bottomrule
 \hline
\end{tabular}
\caption{\label{tab:calibration_weights} Calibration weights for all five iterations of DASH mapping from finger joint angles to motor angles. }
\vspace{-.3cm}
\end{table}

\subsubsection{Teleoperation System}
\label{sec:setup_teleop}

\begin{figure}
    \centering
    \includegraphics[width=0.75\linewidth]{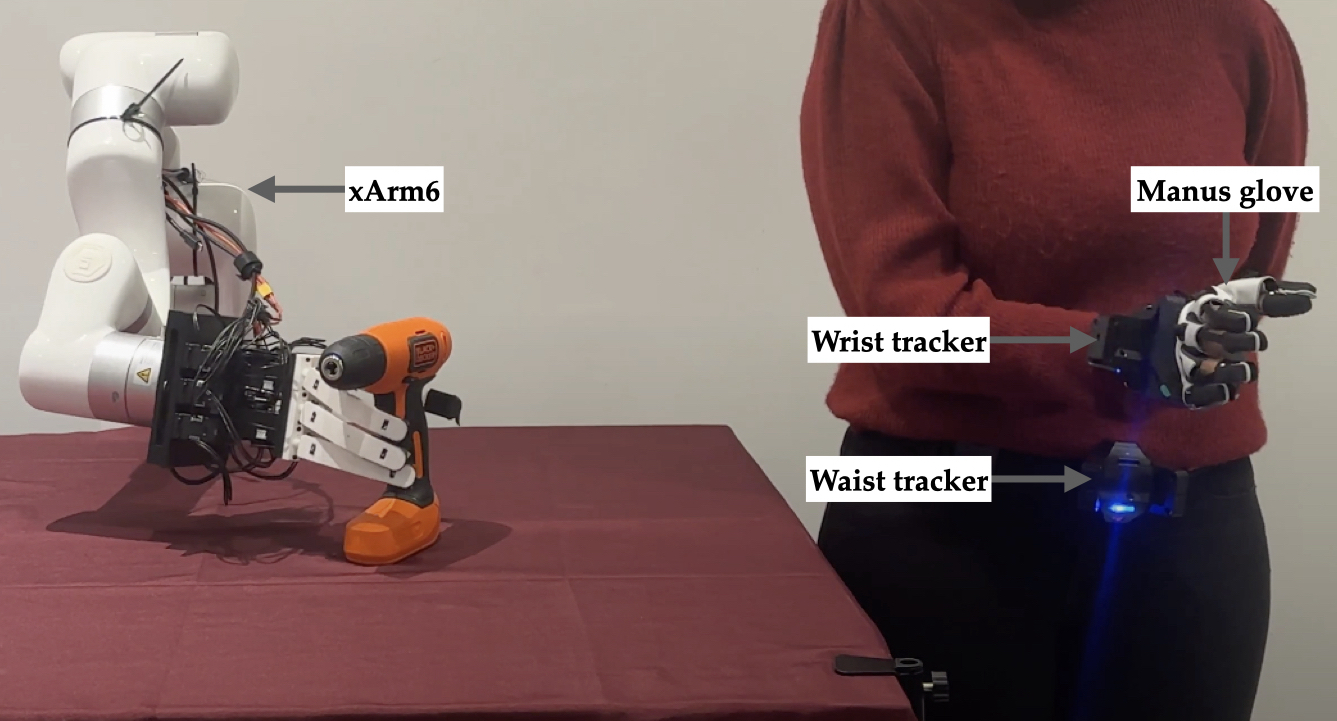}
    \caption{Manus Meta Quantum Metagloves used for tracking the hand for teleoperating the robot arm and \dash{}.}
    \vspace{-.7cm}
    \label{fig:teleop}
\end{figure}

We use Manus Meta Quantum Metagloves~\cite{manus} designed for VR tracking and Mocap Use, as shown in Figure~\ref{fig:teleop} (costs $\sim\$8000$).  The Manus glove is worn on the operator's hand and tracks fingertip positions within a $0.1$-degree accuracy using hall effect sensors.  Each finger returns $4$ angles $\theta_{\mathsf{MCP_{side}}}, \theta_{\mathsf{MCP_{fwd}}}, \theta_{\mathsf{PIP}}, \theta_{\mathsf{DIP}}$ in real time, which are mapped one-to-one to the robot hand. Then, we convert to motor angles using our kinematics models.

To control the robot arm shown in Figure~\ref{fig:teleop}, we employ wearable SteamVR trackers~\cite{steamvr}, utilizing time-of-flight lasers emitted from SteamVR Lighthouses positioned around and above the operator. One tracker is worn on the glove, while another is placed around the waist. We align the waist tracker's rotation with the robot's base frame and adjust the end-effector pose to match the orientation of the human wrist. Then, the human wrist poses are scaled up to cover the robot's larger workspace, making necessary adjustments to ensure user comfort. 
Safety checks, including dynamic force feedback on the arm, prevent accidental damage to the robot or its surrounding environment.
\subsection{Manipulation Tasks}
\label{sec:setup_tasks}
Each \dash{} iteration is tested on a suite of tasks, named \dash{}-30, listed in Table~\ref{tab:tasks} which were inspired by the different types of grasps defined in Liu et al.~\cite{taxonomy} and tasks from previous teleoperation works~\cite{dexpilot, sivakumar2022robotic}. These tasks are categorized by the type of grasp or force necessary. Categories like Hold include a greater number of tasks aimed at testing different grasping techniques and objects. On the other hand, skills like Lever or Twist involve fewer tasks specifically designed to assess whether a particular hand design can successfully perform these skills.
Additionally, some tasks were hand-picked as tasks where compliance of the hand may be advantageous. 

The feedback from the manipulation task evaluation combines observations from the following metrics: task success, performance across five repetitions of the same task, trends in tasks the hand fails to complete, type of grasps possible or used, opposability of fingers, and reachability of fingertips.

\begin{table}[th!]
\centering
\small
\begin{tabular}{@{} *5l @{}}    \toprule
\emph{\textbf{Hold}} \\ \midrule
        1) Scissor,
        2) Hammer,
        3) Chopsticks (single), 
         4) Pen, \\ 
        5) Wooden cylinder (using adduction/abduction),  \\
        6) Screwdriver, 
      7) Drill, 
        8) (Plastic) Egg*, \\
      9) (Plastic) Chip*, 
        10) M\&M* \\ 
        \midrule
\emph{\textbf{Pick (and place)}} \\ \midrule
        11) Dry-Erase Board Eraser, 
        12) Tennis Ball, \\
        13) Softball, 
        14) Cloth*, 
        15) Plush Broccoli, \\ 
      16) Plush Dinosaur,
        17) Pringles Can, 
         18) Spam Box, \\
        19) Mustard Bottle, 
         20) Wine Glass, 
	 21) Bin picking  \\
\midrule  \emph{\textbf{Lever}} \\ \midrule
        22) Cube flip, 
        23) Card pickup from deck \\
\midrule  \emph{\textbf{Twist}} \\ \midrule
  24) Dice rotation in-hand, 
  25) Grape off of stem*  \\ 
\midrule  \emph{\textbf{Open}} \\ \midrule
  26) Plastic bag*, 
  27) Drawer  \\
\midrule  \emph{\textbf{Put in/on}} \\ \midrule
    28) Cup Pouring (onto plate), \\
    29) Cup Stacking \& unstacking, 
    30) 1 inch Block stacking  \\ 
\bottomrule
 \hline
\end{tabular}
\caption{\label{tab:tasks} \dash{}-30: task set of 30 manipulation experiments. Tasks with the asterisk (*) were hand-picked as tasks where compliance of the hand may be advantageous.}
\vspace{-.5cm}
\end{table}


\section{DASH Iterative Design Studies}
\label{sec:DASHdesignandexpts}

\begin{figure*}
    \centering
    \includegraphics[width=0.9\textwidth]{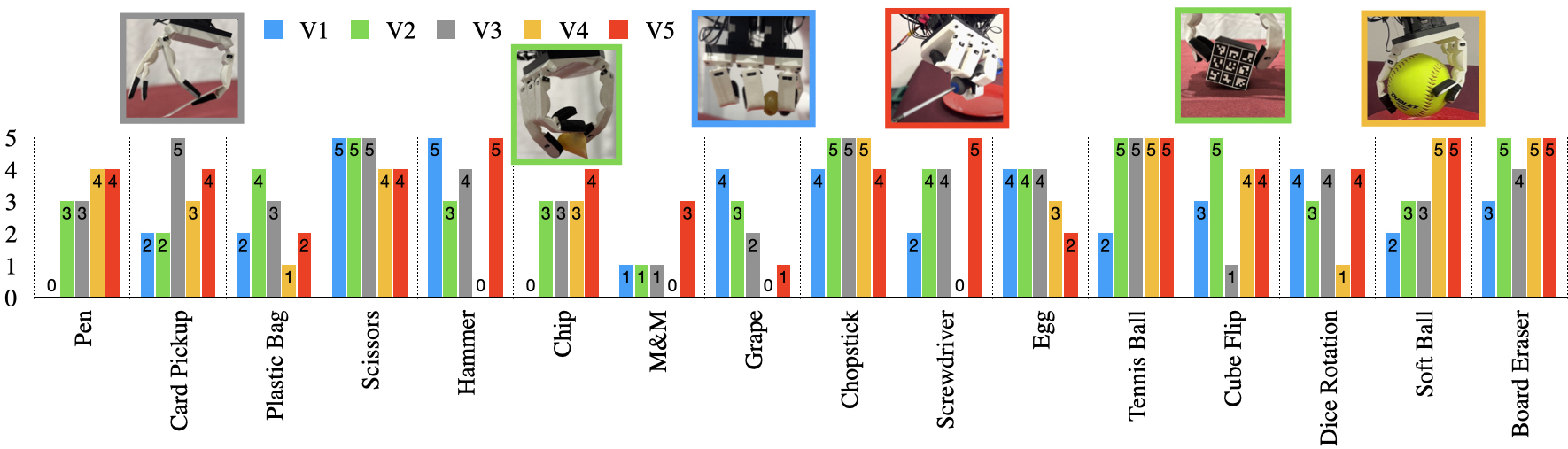}
    \vspace{-.3cm}
    \caption{Subset of tasks with different performance success across \vone{} to \vfive{} on specific tasks used to inform design iteration. The top row of inset images shows representative tasks of successful tasks for each hand.} 
    \label{fig:dash_comparison}
    \vspace{-.5cm}
\end{figure*}

\subsection{Iteration \vone{}}
\label{sec:v1}
\dashvone{} performs best on pick and place tasks and tasks using adduction-abduction such as the grape task, shown in Figure~\ref{fig:dash_comparison} due to the larger hand size and space between fingers as illustrated in Figure~\ref{fig:framework_and_CAD}(b). 
\subsubsection{Design \& Fabrication}
We start by designing the shape of the finger to enable curling fully into itself, incorporating four total tendons per finger, and redesigning the MCP joint as a multi-axis flexure for increased dexterity in our desired hand. 
We iterated on the finger design for approximately four months and use this design for all four fingers in \dashvone{}.

Designing \vone{} included considerations such as tendon anchors, 3D-printing settings, and material stiffness. For example, printing the hand with more infill makes it stiffer but requires more torque than our motors can supply for the joint's full range of motion. To better understand the stiffness of the fingers, we test the finger strength by curling the finger completely and using a force gauge to pull on the finger until it uncurls (see Table~\ref{tab:handiterations} for results). \dashvone{} hand design is shown on the left in Figure~\ref{fig:framework_and_CAD}(b). We designed the full hand assembly for \vone{} in $1$ month.

\subsubsection{Evaluation}
We test \dashvone{} on the 30 manipulation tasks from Section~\ref{sec:setup_tasks}, repeating each task five times. Over 150 repetitions, \dashvone{} succeeded on all 5 repetitions for 10 of the 30 tasks, as shown in Figure~\ref{fig:data_graph}. For \vone{}, these tasks were Scissors, Hammer, Wooden cylinder, Cloth, Plush Broccoli, Plush Dinosaur, Pringles can, Mustard bottle, Wine glass, and Cup stacking. \vone{} struggled to grasp small objects such as the M\&M, Pen, and Chip since the fingers were not able to reach and properly oppose each other. For tasks involving precise motions such as picking the Grape off a stem and opening the Plastic bag, \vone{} uses the abduction-adduction capability.
The abduction-adduction grip strength of \vone{} is high and enables picking up objects such as the grape off the stem with ease, as shown in Figure~\ref{fig:dash_comparison} inset.

\vone{} succeeds on five out of five repetitions on $10$ tasks but shows room for improvement. Grasps that require all four fingers such as picking up a tennis ball would be more successful if the thumb could reach and oppose the rest of the fingers. 
The best opposability to the thumb was to the ring finger, hence pinch (or precision) grasps were easiest to execute with those two fingers. Improving the reachability and opposability of the fingertips requires a smaller palm or longer fingers. We explore these design options in \vtwo{} in order to have more overlap in the workspace of the fingers. 


\subsection{Iteration \vtwo{}}
\label{sec:v2}
\dashvtwo{} performance improves on pick and place and hold tasks requiring power grasps. Furthermore, \vtwo{} excels at the levering task of cube flip on table, shown in Figure~\ref{fig:dash_comparison} inset, due to higher finger strength and fingertip reachability from a smaller palm and longer finger hand design shown in Table~\ref{tab:handiterations} and Figure~\ref{fig:framework_and_CAD}(b).

\subsubsection{Design \& Fabrication}
The second iteration of \dash{} consists of changes to the size of the hand and the MCP joint of the finger. To allow for more reachability among the fingers as well as opposability, the fingers were made longer and the palm was made smaller as shown in Figure~\ref{fig:framework_and_CAD}(b). 
For comparison, \dashvtwo{} is similar in size to the average male hand which is $88.9$mm wide and $193$mm long (wrist to fingertip)~\cite{handsize}~. 
Compared to \dashvone{}, there is more than a 25\% reduction in area of the palm and the finger length increased by 11\% in \vtwo{} which is shown in Figure~\ref{fig:framework_and_CAD}(b).

The MCP joint was improved to achieve a larger range of motion. 
The underlying structure of the MCP joint is a cylinder to act as a multi-axis flexure, thus we increase the height of the cylinder to increase the joint angle range for the side-to-side and forward motion of the fingers. 
The design changes also resulted in a higher maximum load of a single finger as shown under finger strength in Table~\ref{tab:handiterations}. 
Thus, \vtwo{} achieves increased range of side-to-side and forward motion for the fingers by redesigning the MCP joint, and has a larger overlap in the workspace of the fingers solving the reachability and opposability issues in \vone{}.


Designing, printing, and assembling \vtwo{} took 5, 83, and 6.5 hours, respectively. 
Printing \vtwo{} required us to not only re-print the soft hand, but also the rigid motor housing as the motor arrangement differs from \vone{}.
In total, making \vtwo{} from \vone{} took 94 hours.

\subsubsection{Evaluation}
With larger range of motion at the MCP joint and better reachability, we expect \vtwo{} to achieve better performance on tasks involving smaller objects like M\&M, Pen, and Chip. As shown in Figure~\ref{fig:dash_comparison}, \vtwo{} did improve performance on Pen and Chip. M\&M and Card pickup were tasks that did not improve from \vone{}. Both of these tasks require fine manipulation which is still a limitation in \vtwo{}. Instead, our main improvement from \vone{} to \vtwo{} is in achieving better power grasps. Tripod grasps or using more than two fingers was necessary to have stable grasps, especially for the holding tasks such as Hammer, Screwdriver, and Chopstick.  
However, observations during teleoperation included difficulty using precision grasps with two fingers. 

\vtwo{} performs better than \vone{} in 14 tasks (refer to Figure~\ref{fig:data_graph}), including tasks involving Soft ball, Screw driver, Tennis ball, Dry-erase board eraser, and Spam box that all require power grasps. As shown in Figure~\ref{fig:dash_comparison}, the most significant improvements are seen for Pen, Chip, Tennis ball and Cube Flip.
The inset in Figure~\ref{fig:dash_comparison} shows \vtwo{} grasping Chip with the ring finger and thumb finger, and \vtwo{} succeeding at all five repetitions of Cube Flip. These improvements are possible with better reachability and opposability of the thumb with the rest of the fingertips. 

Having more space between the fingers made abduction-adduction tasks such as picking Grape off of a stem and Wooden cylinder easier for \vone{} compared to \vtwo{}, but \vtwo{} still performs reasonably well. Out of the 150 repetitions, \vtwo{} is successful in 123 repetitions, which is 18 more when compared to \vone{}. Additionally, the number of tasks where all five repetitions were successful increased from 10 tasks using \vone{} to 14 tasks using \vtwo{}. 

\begin{table}[t]
\centering
\begin{tabular}{@{} *7l @{}}    \toprule
\emph{\textbf{Hand design}} & \emph{v1} & \emph{v2} & \emph{v3} & \emph{v4} & \emph{v5} \\ \midrule
Palm size & 94x102 & 84x84 & 84x84 & 84x84 & 84x84 \\
Finger length & 90 & 100 & 100 & 100 & 100 \\ 
MCP diameter & 6 & 6  & 6  &  10 &  8 \\
MCP height  & 6 & 8 & 8 & 8 & 8\\
DIP crease width & 10.3 & 10.3 & 8.9 & 10.3 & 13.0 \\
Thumb angle & 45$^\circ$ & 45$^\circ$ & 0$^\circ$ & 22.5$^\circ$ & 22.5$^\circ$\\
Fingertip edge & 3.5 & 3.5 & 1.73 & 3.5 & 3.5 \\
Fingertip thickness & 13.21 & 13.22 & 7.98 & 11.22 & 8.75 \\
Finger strength & 37.8 & 47.6 & 34.5 & 51.8 & 27.4 \\
\bottomrule
 \hline
\end{tabular}
\caption{\label{tab:handiterations} Hand design parameters where finger length refers to the distances in millimeters from the top of the MCP joint to the fingertip and finger strength (N) is measured by pulling on a fully curled finger with a digital force gauge.}
\vspace{-.7cm}
\end{table}

\subsection{Iteration \vthree{}}
\label{sec:v3}
\dashvthree{} has the best thumb opposability and thinnest fingertip design out of all of our hand iterations, yielding in the best score for Card Pickup as shown in Figure~\ref{fig:dash_comparison}. Thinner fingertips, however, led to weaker finger strength which decreased task success for tasks such as Dry-erase Board Eraser and Grape off stem.
\subsubsection{Design \& Fabrication}
The changes from \dashvtwo{} to \vthree{} involve changing the thumb placement and fingertip shape. In order to make grasps with only two or three fingers more stable, the thumb has to be directly opposable to the rest of the fingers, most importantly the index finger. In \vtwo{}, the thumb has a 45-degree angle to the palm which we change to be parallel to the index finger in \vthree{}, as shown in the middle of Figure~\ref{fig:framework_and_CAD}(b) and in Table~\ref{tab:handiterations}. 
In addition to the thumb placement, the fingertip shape was changed from a rounded surface to a thinner wedge-like surface (see Figure~\ref{fig:framework_and_CAD}). 
The rounded surface in \vtwo{} presented a point contact when interacting with objects. In contrast, the wedge-like surface will have a larger contact area and thinner fingertip (similar to fingernail) in order to get under objects to grasp. This results in a thin fingertip edge, almost half the size of \vone{} and \vtwo{}'s fingertip edge (see Table~\ref{tab:handiterations}).
We also move the tendon routing farther away from the center axis of the MCP joint so that we can exert more torque when folding the finger forward about the MCP joint. 

Designing, printing, and assembling \vthree{} took 4, 67.25, and 4.25 hours, respectively. Similar to \vtwo{}, we reprinted the motor housing again due to the new thumb placement. In total, making \vthree{} took almost 83.75 hours.

\subsubsection{Evaluation}
\begin{figure}
    \centering
    \includegraphics[width=\linewidth]{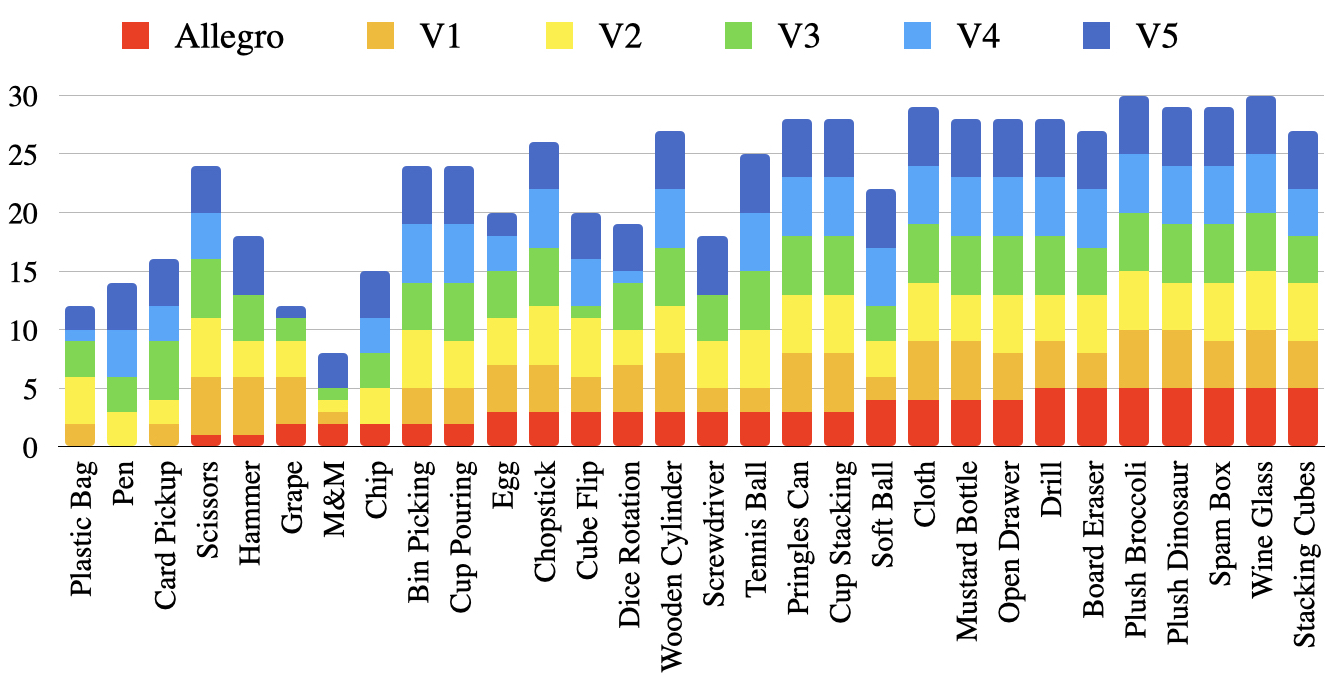}
    \vspace{-0.3cm}
    \caption{Task performance over 5 repetitions of each task across \vone{}, \vtwo{}, \vthree{}, \vfour{}, \vfive{}, and Allegro as baseline. The tasks are ordered difficult to easy from left to right, according to task performance of Allegro.}
    \vspace{-0.7cm}
    \label{fig:data_graph}
\end{figure}

As shown in Figure~\ref{fig:data_graph}, \dashvthree{} has more successful tasks than the previous hand iterations and our baseline, completing 16 tasks successfully in all repetitions as opposed to the 14 tasks \vtwo{} successfully executed. \vthree{} succeeded on all repetitions of Wooden Cylinder, Card Pickup, Cup Pouring, Drill, Plush Dinosaur, and Mustard Bottle, which are tasks \vtwo{} did not master.  The task improvement was due to better thumb opposability compared to \vtwo{}. In total, \vthree{} succeeded on 124 repetitions which is 1 more than the number of repetitions \vtwo{} is successful at. 
With \vthree{}, we observe higher grasp stability during power grasps and handling of delicate objects, during teleoperation. Additionally, we find that the fingertip shape makes a large difference for specific tasks. We clearly see this effect occurring in Cube flip and Card pickup (see inset images of \vtwo{} Cube Flip and \vthree{} Card Pickup in Figure~\ref{fig:dash_comparison}). The flat fingertips of \vthree{} are ideal for thin delicate card pickup but not for the cube flip. Reorienting the cube in-hand in Cube flip is better suited to the rounded fingertip on \vtwo{}, keeping a stable point contact while the object rotates on the table. 



\subsection{Iteration \vfour{}}
\label{sec:v4}
\dashvfour{} was optimized for strength as we found that lacking for tasks such as Cube Flip for \vthree{}. This allowed for heavy objects like Soft Ball to have great success with \vfour{} as shown in Figure~\ref{fig:dash_comparison} but decreased finger folding motion resulted in decreased performance for tasks such as Hammer, Screwdriver, and M\&M. 
\subsubsection{Design \& Fabrication}
The fourth iteration of \dash{} was designed to optimize for strength. We focused on redesigning the MCP joint to be thicker, providing increased stiffness for folding the fingers into the palm. Achieving the right balance was challenging, as we aimed to maintain the range of motion for MCP forward motion within the torque limits of our motors. While a simple solution would be to use larger motors to increase force and stiffness at the MCP joint, this would result in a larger and heavier hand. 

Additionally, we made changes to the fingertips and thumb placement, creating a hybrid design influenced by \dashvtwo{} and \dashvthree{}. Thicker fingertips proved useful for tasks involving rotation, such as Cube flipping, while thinner fingertips were beneficial for pinch grasps like Card pickup. The result was rounded edges with a flat surface in the center of the fingertip, providing versatility for pinch to power grasps. Similarly, the thumb placement was positioned between \vtwo{} at $45^\circ$ and \vthree{} at $0^\circ$, settling at $22.5^\circ$ relative to the palm. While \vtwo{} excelled in power grasps and \vthree{} in pinch grasps like Card pickup, we aimed for \vfour{} to perform equally well in both types of grasping.

Designing, printing, and assembling \vfour{} took 8.5, 82, and 5 hours, respectively. Similar to \vthree{}, we reprinted the motor housing to accommodate the new thumb placement.

\subsubsection{Evaluation}
\dashvfour{} successfully completed all five repetitions of 17 tasks, surpassing the task performance of \vthree{}. \vfour{} maintained its performance in most of these tasks, with the exception of Scissors, as shown in Figure~\ref{fig:dash_comparison}. However, it outperformed \vthree{} in tasks involving the Dry-erase board eraser and performed better than any previous hand iteration in the Soft ball task. This was attributed to the stronger MCP joint, which enhanced the finger strength, as indicated in Table~\ref{tab:handiterations}. Nevertheless, the limited range of motion in the MCP forward joint resulted in poor reachability, causing objects like Scissors to slip between the fingertips.

Overall, the hybrid thumb position and fingertip shape, combining features from \vtwo{} and \vthree{}, proved advantageous in achieving a greater number of tasks. However, the next iteration should address the loss of range of forward folding motion to improve reachability. The limited reachability of \vfour{} also led to zero successes out of five repetitions in four tasks, including Hammer, Screwdriver, M\&M, and Grape off stem. All of these limitations can be attributed to the restricted range of motion in the MCP forward joint.

\subsection{Iteration \vfive{}}
\label{sec:v5}
\dashvfive{} aimed to be a combination of all previous hand design features with respect to joint and fingertip thicknesses. \vfive{} generally outperformed all previous design iterations and excelled at the Screwdriver task as shown in Figure~\ref{fig:first_fig}.
\subsubsection{Design \& Fabrication}
The fifth iteration of \dash{} features a stiffer MCP joint compared to \vthree{}, but it is more compliant than the MCP joint of \vfour{}. By increasing the compliance at the MCP joint, we were able to achieve a greater range of motion at the joint compared to \vfour{}, which had limited folding capabilities. Furthermore, we made the fingertip thinner than that of \vfour{}, and widened the DIP crease (as shown in Table~\ref{tab:handiterations}), in order to improve the curling of the finger. As a result, \dashvfive{} exhibits the most extensive curling motion among all the previous iterations.

Designing, printing, and assembling \vfive{} took 2, 24, and 2.75 hours, respectively. Unlike the previous versions, we kept the motor assembly unchanged and only replaced the fingers of \vfour{}. Consequently, the total time required for iteration was the lowest for \vfive{}, totaling 28.75 hours.
\subsubsection{Evaluation}
Among all the design iterations of \dash{}, \dashvfive{} performed the best. \vfive{} succeeded on five out of five repetitions on $19$ tasks and achieved a completion rate of $131$ out of $150$ total task repetitions. In addition to the tasks that \vfour{} succeeded on, \vfive{} also completed five out of five repetitions on the Hammer and Stacking cubes tasks. This improvement indicates that we have made incremental progress on the hand design. \vfive{} had the most curling range of motion than previous hands which made picking objects easier for the teleoperating user due to stable grasps enveloping objects into the palm.

As shown in Figure~\ref{fig:dash_comparison}, \vfive{} showed improved task performance for Hammer, Screwdriver, Chip, M\&M, Grape off stem, and Plastic Bag. However, it performed worse for the Chopsticks and Egg tasks. Although \vfive{} has the lowest finger strength among all \dash{} iterations due to thinner joints and thinner fingertips (as shown in Table~\ref{tab:handiterations}), its enhanced finger curling abilities even enabled a single finger to hold objects. However, when completely curled, thin objects such as the chopsticks were prone to falling between the thumb and fingers. This issue could be addressed by introducing longer fingers to allow for more overlap between the fingertips. Overall, \vfive{} outperformed all previous iterations of \dash{} across all 30 tasks. However, it is worth noting that certain hands may specialize in specific tasks. For instance, \vfive{} excelled at picking up Screwdriver, while \vthree{} was the most suitable for Card pickup, as shown in Figure~\ref{fig:dash_comparison}. One interesting result involved the \vfive{} screwdriver hold, which aligned perfectly in the groove on the tool handle.

\begin{figure}
    \centering
    \includegraphics[width=\linewidth]{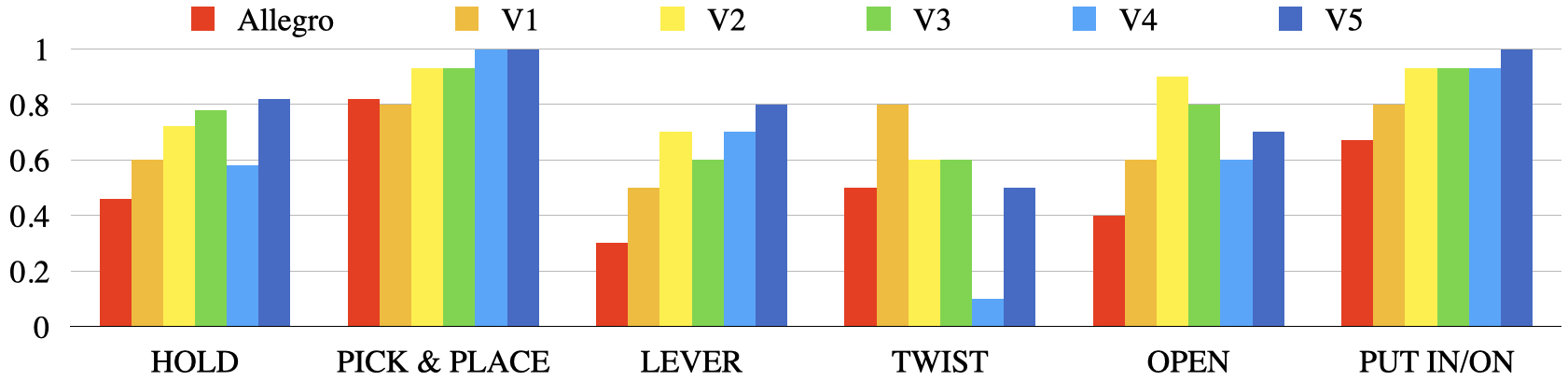}
    \caption{Task performance across \vone{}, \vtwo{}, \vthree{}, \vfour{}, \vfive{}, and Allegro as baseline on each category of tasks from Table~\ref{tab:tasks}. }
    \vspace{-0.7cm}
    \label{fig:category_graph}
\end{figure}

\section{Baseline Study: Allegro Dexterous Hand}
\label{sec:allegro}
Allegro~\cite{lee2016kitech} is an off-the-shelf gripper that we use as a baseline. Allegro is a dexterous robotic hand that has four fingers with motors at the joints, rigid structure, and large rubber spherical fingertips. We perform the same 30 manipulation tasks from \dash{}-30 (Table~\ref{tab:tasks}) with Allegro to compare the performance against all iterations of DASH. 

Allegro succeeded on all five repetitions on 7 out of 30 tasks. These tasks included manipulating the Drill, Dry-Erase board eraser, Plush broccoli, Plush Dinosaur, Spam box, Wine glass, and Stacking cubes, as shown as the rightmost tasks in Figure~\ref{fig:data_graph}. 
Allegro performed best on pick and place tasks compared to other types of tasks as shown in Figure~\ref{fig:category_graph}. However, all iterations of \dash{}, except \vone{}, were also successful at these tasks.

The Allegro robotic hand and fingers had difficulty with tasks such as picking up the Pen, Card, Plastic bag, Scissors, and Hammer which required precision.
While both \dash{} and Allegro hands lack sensing capabilities, this disproportionately affected Allegro because lack of compliance made it easy to grasp too tightly or not enough, especially for rigid objects.
Similarly, the Cup pouring grasp was unstable due to the spherical fingertips rotating the cup in-hand during the task. The side-to-side motion (or abduction-adduction) of the fingers was limited, making Dice Rotation coarse and unpredictable.  However, Allegro had stable grasps for larger and softer objects such as the Drill, Softball, Plush Dinosaur, Plush Broccoli, and Wine Glass (see rightmost tasks in Figure~\ref{fig:data_graph}).
 
\section{Discussion}
\label{sec:Discussion}
Across the $30$ tasks, we observe that \vfive{} has the best performance solving $19$ tasks successfully completing all repetitions, while \vfour{}, \vthree{}, \vtwo{}, \vone{}, and Allegro solve $17$, $16$, $14$, $10$, and $7$ tasks, respectively. In Figure~\ref{fig:category_graph}, we see that all iterations of \dash{} outperform the Allegro baseline on most categories of tasks listed in Table~\ref{tab:tasks} as well as steady improvement in \dash{} iterations except for twisting and opening which are the most difficult categories of tasks. The two twisting tasks were Dice Rotation and Grape which both more successful with the larger palm and space between fingers for \vone{} compared to other iterations. For opening tasks, \vtwo{} had more success on opening Plastic Bag due to its rounder fingertips and higher finger strength. Through our suite of varied manipulation tasks and human-in-the-loop design iteration, we validate our framework's ability to use real world evaluation to iteratively design soft robot hands through rapid prototyping and teleoperation. 

From our case study in Section~\ref{sec:DASHdesignandexpts}, we draw three crucial observations regarding our proposed framework. Firstly, the direct feedback from the designer performing real world manipulation tasks with \dash{} was crucial for us in informing the design changes required to improve performance across iterations. In contrast, testing in simulation can result in design changes that do not necessarily translate to performance improvement in the real world. Secondly, using teleoperation removed the necessity of designing different control policies for $30$ various tasks across six robot hand morphologies in our case study, and allowed us to adjust grasps in real-time during task execution, which is often not feasible in simulation or by using keyframed poses. Lastly, despite using real robot hands in the design iteration process, our framework has a short iteration time, consisting mostly of printing time (about $80\%$ of total time), by leveraging 3D-printing and the use of teleoperation to evaluate the design in the real world. 

Our framework can extend to testing other soft robotic hands in the real world for rapid design iteration. There are three stages of our framework, as shown in Figure~\ref{fig:framework_and_CAD}(a), including design, fabrication, and evaluation. Some best practices include incorporating a modular design to facilitate easier iteration, adopting rapid prototyping methods for seamless fabrication, and favoring incremental design changes to allow for targeted iteration on specific design features. Our method of evaluating using teleoperation also allowed for minimal changes in control when the hand design changed. Our framework can be used to test easily prototyped hands, such as those by Bauer et al.~\cite{bauer2022towards} or RBO~\cite{Puhlmann_2022}, using the same setup used for \dash{} iteration, similar to our Manus~\cite{manus} VR teleoperation system. Additionally, \dash{}-30, our suite of 30 varied manipulation tasks can be used to benchmark other dexterous hands in the community. 

Observing that our robot hand has similar structure and size to human hands, we note a crucial limitation of our framework, shown in Figure~\ref{fig:framework_and_CAD}(a), for robot hand morphologies that diverge from human hand morphology as teleoperation might not be feasible in such cases. Additionally, calibration or mapping of the teleoperator's hand to the robot hand can have a significant impact on the robot hand's performance in real-world manipulation tasks. For example, an inaccurate mapping from the teleoperator's hand to the robot hand can incorrectly evaluate the robot hand to be incapable of some tasks. Another limitation for this framework is that it can result in longer turnover times for designs that cannot be made with rapid prototyping techniques such as 3D-printing. Lastly, monotonic improvement is difficult to guarantee due to the manual design iteration process in our framework.


\section{Conclusion and Future Work}
\label{sec:Conclusion_FW}
This paper presents a design iteration process that can supplement existing design iteration techniques by leveraging 3D-printing and teleoperation. 
We exhibit the potential of this framework through a case study of designing a 16-DoF 3D-printed dexterous anthropomorphic soft hand \dash{}. By 3D-printing the new design at each iteration, and evaluating it on real-world manipulation tasks using teleoperation to inform future hand designs, we consistently improve its performance over the baseline Allegro hand and across successive iterations of \dash{}. We open-sourced our \dash{} CAD models and teleoperated demonstration data at \url{https://dash-through-interaction.github.io}.

Future directions include automatic design iteration by singling out features of the CAD design and correlating them with capabilities of the hand.  Further study would be required to automate this process and use collected data to learn what properties of the hand should be improved for better task performance. 
Currently, the process of design iteration in our case study was manual in that we chose parameters to change based on task performance and observations from real-world manipulation experiments.  






\bibliographystyle{IEEEtran}
\bibliography{IEEEabrv,references}

\begin{thebibliography}{10}
\providecommand{\url}[1]{#1}
\csname url@rmstyle\endcsname
\providecommand{\newblock}{\relax}
\providecommand{\bibinfo}[2]{#2}
\providecommand\BIBentrySTDinterwordspacing{\spaceskip=0pt\relax}
\providecommand\BIBentryALTinterwordstretchfactor{4}
\providecommand\BIBentryALTinterwordspacing{\spaceskip=\fontdimen2\font plus
\BIBentryALTinterwordstretchfactor\fontdimen3\font minus
  \fontdimen4\font\relax}
\providecommand\BIBforeignlanguage[2]{{%
\expandafter\ifx\csname l@#1\endcsname\relax
\typeout{** WARNING: IEEEtran.bst: No hyphenation pattern has been}%
\typeout{** loaded for the language `#1'. Using the pattern for}%
\typeout{** the default language instead.}%
\else
\language=\csname l@#1\endcsname
\fi
#2}}

\bibitem{chen2020design}
F.~Chen and M.~Y. Wang, ``Design optimization of soft robots: A review of the
  state of the art,'' \emph{IEEE Robotics \& Automation Magazine}, vol.~27,
  no.~4, pp. 27--43, 2020.

\bibitem{della2018toward}
C.~Della~Santina, C.~Piazza, G.~Grioli, M.~G. Catalano, and A.~Bicchi, ``Toward
  dexterous manipulation with augmented adaptive synergies: The pisa/iit
  softhand 2,'' \emph{IEEE Transactions on Robotics}, vol.~34, no.~5, pp.
  1141--1156, 2018.

\bibitem{Puhlmann_2022}
\BIBentryALTinterwordspacing
S.~Puhlmann, J.~Harris, and O.~Brock, ``{RBO} hand 3: A platform for soft
  dexterous manipulation,'' \emph{{IEEE} Transactions on Robotics}, vol.~38,
  no.~6, pp. 3434--3449, dec 2022. [Online]. Available:
  \url{https://doi.org/10.1109%2Ftro.2022.3156806}
\BIBentrySTDinterwordspacing

\bibitem{bauer2022towards}
D.~Bauer, C.~Bauer, A.~Lakshmipathy, R.~Shu, and N.~S. Pollard, ``Towards very
  low-cost iterative prototyping for fully printable dexterous soft robotic
  hands,'' in \emph{2022 IEEE 5th International Conference on Soft Robotics
  (RoboSoft)}.\hskip 1em plus 0.5em minus 0.4em\relax IEEE, 2022, pp. 490--497.

\bibitem{zhang2022creation}
Y.~Zhang and R.~K. Katzschmann, ``Creation of a modular soft robotic fish
  testing platform,'' \emph{arXiv preprint arXiv:2201.04098}, 2022.

\bibitem{abondance2020dexterous}
S.~Abondance, C.~B. Teeple, and R.~J. Wood, ``A dexterous soft robotic hand for
  delicate in-hand manipulation,'' \emph{IEEE Robotics and Automation Letters},
  vol.~5, no.~4, pp. 5502--5509, 2020.

\bibitem{FrankMocap_2021_ICCV}
Y.~Rong, T.~Shiratori, and H.~Joo, ``Frankmocap: A monocular 3d whole-body pose
  estimation system via regression and integration,'' in \emph{Proceedings of
  the IEEE/CVF International Conference on Computer Vision (ICCV) Workshops},
  October 2021, pp. 1749--1759.

\bibitem{sivakumar2022robotic}
A.~Sivakumar, K.~Shaw, and D.~Pathak, ``Robotic telekinesis: learning a robotic
  hand imitator by watching humans on youtube,'' \emph{arXiv preprint
  arXiv:2202.10448}, 2022.

\bibitem{lee2016kitech}
D.-H. Lee, J.-H. Park, S.-W. Park, M.-H. Baeg, and J.-H. Bae, ``Kitech-hand: A
  highly dexterous and modularized robotic hand,'' \emph{IEEE/ASME Transactions
  on Mechatronics}, vol.~22, no.~2, pp. 876--887, 2016.

\bibitem{Controzzi2014}
\BIBentryALTinterwordspacing
M.~Controzzi, C.~Cipriani, and M.~C. Carrozza, \emph{Design of Artificial
  Hands: A Review}.\hskip 1em plus 0.5em minus 0.4em\relax Cham: Springer
  International Publishing, 2014, pp. 219--246. [Online]. Available:
  \url{https://doi.org/10.1007/978-3-319-03017-3_11}
\BIBentrySTDinterwordspacing

\bibitem{feng2018soft}
N.~Feng, Q.~Shi, H.~Wang, J.~Gong, C.~Liu, and Z.~Lu, ``A soft robotic hand:
  design, analysis, semg control, and experiment,'' \emph{The International
  Journal of Advanced Manufacturing Technology}, vol.~97, pp. 319--333, 2018.

\bibitem{elsayed2014finite}
Y.~Elsayed, A.~Vincensi, C.~Lekakou, T.~Geng, C.~Saaj, T.~Ranzani,
  M.~Cianchetti, and A.~Menciassi, ``Finite element analysis and design
  optimization of a pneumatically actuating silicone module for robotic surgery
  applications,'' \emph{Soft Robotics}, vol.~1, no.~4, pp. 255--262, 2014.

\bibitem{deimel2017automated}
R.~Deimel, P.~Irmisch, V.~Wall, and O.~Brock, ``Automated co-design of soft
  hand morphology and control strategy for grasping,'' in \emph{2017 IEEE/RSJ
  International Conference on Intelligent Robots and Systems (IROS)}.\hskip 1em
  plus 0.5em minus 0.4em\relax IEEE, 2017, pp. 1213--1218.

\bibitem{nagabandi2020deep}
A.~Nagabandi, K.~Konolige, S.~Levine, and V.~Kumar, ``Deep dynamics models for
  learning dexterous manipulation,'' in \emph{Conference on Robot
  Learning}.\hskip 1em plus 0.5em minus 0.4em\relax PMLR, 2020, pp. 1101--1112.

\bibitem{andrychowicz2020learning}
O.~M. Andrychowicz, B.~Baker, M.~Chociej, R.~Jozefowicz, B.~McGrew,
  J.~Pachocki, A.~Petron, M.~Plappert, G.~Powell, A.~Ray, \emph{et~al.},
  ``Learning dexterous in-hand manipulation,'' \emph{The International Journal
  of Robotics Research}, vol.~39, no.~1, pp. 3--20, 2020.

\bibitem{dime_lerrel}
\BIBentryALTinterwordspacing
S.~P. Arunachalam, S.~Silwal, B.~Evans, and L.~Pinto, ``Dexterous imitation
  made easy: A learning-based framework for efficient dexterous manipulation,''
  2022. [Online]. Available: \url{https://arxiv.org/abs/2203.13251}
\BIBentrySTDinterwordspacing

\bibitem{videodex}
K.~Shaw, S.~Bahl, and D.~Pathak, ``Videodex: Learning dexterity from internet
  videos,'' in \emph{Conference on Robot Learning (CoRL)}, 2022.

\bibitem{li2022survey}
R.~Li, H.~Wang, and Z.~Liu, ``Survey on mapping human hand motion to robotic
  hands for teleoperation,'' \emph{IEEE Transactions on Circuits and Systems
  for Video Technology}, vol.~32, no.~5, pp. 2647--2665, 2022.

\bibitem{liarokapis2013telemanipulation}
M.~V. Liarokapis, P.~K. Artemiadis, and K.~J. Kyriakopoulos, ``Telemanipulation
  with the dlr/hit ii robot hand using a dataglove and a low cost force
  feedback device,'' in \emph{21st Mediterranean Conference on Control and
  Automation}.\hskip 1em plus 0.5em minus 0.4em\relax IEEE, 2013, pp. 431--436.

\bibitem{ninjaflex}
``Ninjaflex edge,''
  \href{https://ninjatek.com/shop/edge/}{https://ninjatek.com/shop/edge/},
  accessed on 2022-11-26.

\bibitem{xarm6}
``Ufactory xarm6,''
  \href{https://www.ufactory.cc/product-page/ufactory-xarm-6}{https://www.ufactory.cc/product-page/ufactory-xarm-6},
  accessed on 2023-02-03.

\bibitem{zhu2019dexterous}
H.~Zhu, A.~Gupta, A.~Rajeswaran, S.~Levine, and V.~Kumar, ``Dexterous
  manipulation with deep reinforcement learning: Efficient, general, and
  low-cost,'' in \emph{2019 International Conference on Robotics and Automation
  (ICRA)}.\hskip 1em plus 0.5em minus 0.4em\relax IEEE, 2019, pp. 3651--3657.

\bibitem{manus}
``Manus,'' \href{https://www.manus-meta.com}{https://www.manus-meta.com},
  note={Accessed on 2022-11-28}.

\bibitem{steamvr}
``Steam vr,'' \href{https://www.steamvr.com/en/}{https://www.steamvr.com/en/},
  accessed on 2023-02-03.

\bibitem{taxonomy}
J.~Liu, F.~Feng, Y.~C. Nakamura, and N.~S. Pollard, ``A taxonomy of everyday
  grasps in action,'' in \emph{2014 IEEE-RAS International Conference on
  Humanoid Robots}, 2014, pp. 573--580.

\bibitem{dexpilot}
\BIBentryALTinterwordspacing
A.~Handa, K.~Van~Wyk, W.~Yang, J.~Liang, Y.-W. Chao, Q.~Wan, S.~Birchfield,
  N.~Ratliff, and D.~Fox, ``Dexpilot: Vision based teleoperation of dexterous
  robotic hand-arm system,'' 2019. [Online]. Available:
  \url{https://arxiv.org/abs/1910.03135}
\BIBentrySTDinterwordspacing

\bibitem{handsize}
``Healthline,''
  \href{https://www.healthline.com/health/average-hand-size#adults}{https://www.healthline.com/health/average-hand-size\#adults},
  accessed on 2022-11-29.

\end{thebibliography}

\end{document}